\documentclass{ecai}

\usepackage{graphicx}
\usepackage{latexsym}

\usepackage{amsmath}
\usepackage{breqn}

\usepackage{lipsum,mwe,cuted}
\usepackage{float}
\usepackage{caption}
\usepackage{stfloats}
\usepackage{titlesec}
\usepackage{subfigure}

\usepackage{pifont}

\usepackage{fontawesome}

\usepackage{hyperref}

\usepackage{booktabs}
\usepackage{graphicx}
\usepackage{subfigure}
\usepackage{algorithm}
\usepackage{algpseudocode}

\usepackage{booktabs}
\usepackage{multirow}
\usepackage[table,xcdraw]{xcolor}

\newcommand*\rot{\rotatebox{90}}

\titlespacing*{\section}{0pt}{*1.0}{*1.0}
\titlespacing*{\subsection}{0pt}{*0.5}{*0.5}
\titlespacing*{\subsubsection}{0pt}{*0.1}{*0.1}
\begin{document}

\begin{frontmatter}

\title{DTBS: Dual-Teacher Bi-directional Self-training for Domain Adaptation in Nighttime Semantic Segmentation}

\author[A]{\fnms{Fanding}~\snm{Huang}}
\author[B]{\fnms{Zihao}~\snm{Yao}}
\author[A]{\fnms{Wenhui}~\snm{Zhou}\thanks{Corresponding Author. Email: zhouwenhui@hdu.edu.cn}} 

\address[A]{Hangzhou Dianzi University, Hangzhou, China}
\address[B]{University of Wollongong, Wollongong, Australia}

\begin{abstract}
Due to the poor illumination and the difficulty in annotating, nighttime conditions pose a significant challenge for autonomous vehicle perception systems. Unsupervised domain adaptation (UDA) has been widely applied to semantic segmentation on such images to adapt models from normal conditions to target nighttime-condition domains. Self-training (ST) is a paradigm in UDA, where a momentum teacher is utilized for pseudo-label prediction, but a confirmation bias issue exists. Because the one-directional knowledge transfer from a single teacher is insufficient to adapt to a large domain shift. To mitigate this issue, we propose to alleviate domain gap by incrementally considering style influence and illumination change. Therefore, we introduce a one-stage Dual-Teacher Bi-directional Self-training (DTBS) framework for smooth knowledge transfer and feedback. Based on two teacher models, we present a novel pipeline to respectively decouple style and illumination shift. In addition, we propose a new Re-weight exponential moving average (EMA) to merge the knowledge of style and illumination factors, and provide feedback to the student model. In this way, our method can be embedded in other UDA methods to enhance their performance. For example, the Cityscapes to ACDC night task yielded 53.8 mIoU (\%), which corresponds to an improvement of +5\% over the previous state-of-the-art. The code is available at \url{https://github.com/hf618/DTBS}.
\end{abstract}

\end{frontmatter}

\section{Introduction}
%

Semantic segmentation is a significant computer vision issue that labels each pixel of an image. It is used in numerous applications, such as autonomous driving, medical image analysis and augmented reality. Although some techniques have been presented \cite{fu2019dual,huang2019ccnet}, they are primarily utilized for training on daytime images with varying illumination. However, deploying these models in real-world conditions, especially in fog, rain, snow, and darkness environments, demands more robust performance.  Existing models trained on available datasets often fail to deliver desirable results. Therefore, there's a pressing need to develop new strategies to learn robust segmentation models that can generalize well under adverse conditions. This paper's primary focus is nighttime semantic segmentation across domains without ground truth (GT), which is a vital challenge in autonomous driving.




\begin{figure}[h]
	\small
	\setlength{\abovecaptionskip}{0.1cm}
	\centering
       
	 \includegraphics[width=\textwidth/2]{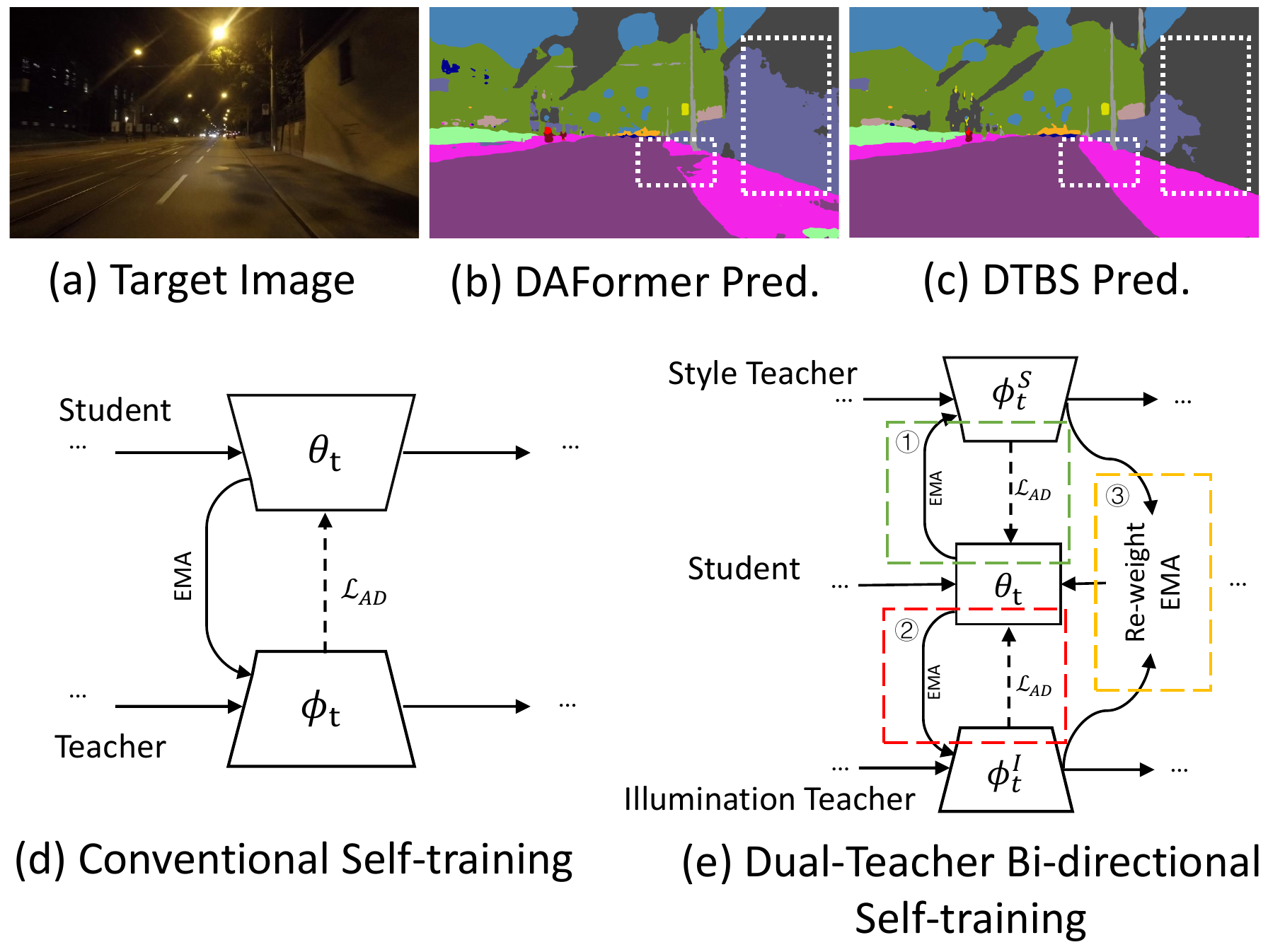}
	\caption{(a) Street scenes in the target domain with indistinguishable neighboring classes. (b) Previous works like DAFormer \cite{hoyer2022daformer} cannot effectively identify confusing classes, e.g. mistakenly predicting sidewalk as road and building as wall. (c) The proposed DTBS framework enhances the learning of different urban street styles, such as the sidewalk and building in the foreground. (d) ST employs a teacher to supervise the student, transferring information from student to teacher. (e) Our DTBS derives style teacher and illumination teacher with style knowledge and illumination knowledge smoothly, respectively, allowing student and teachers to learn from each other.} 
        \label{intro}
  
\end{figure}

To tackle this issue, researchers proposed unsupervised domain adaptation (UDA) techniques aimed at reducing the domain shift between the source and target domains. In \cite{yang2020fda,reinhard2001color,zhu2017unpaired}, for instance, the image translation network generates synthetic datasets by stylizing the original images. Nonetheless, these methods cannot effectively exploit the semantic embedding of the segmentation task, which ignores the influence of pixel-wise label correspondence. Specifically, when converting daytime images to nighttime images, FDA \cite{yang2020fda} or Color Transfer \cite{reinhard2001color} may result in artifacts that lead to severe negative transfer. In severe cases, several illumination sources (traffic lights, street lights, etc.) may cause local overexposure to brightness. 


Some researchers \cite{dai2018dark,sakaridis2019guided,sakaridis2020map} used twilight images as a bridge, which have the same GPS location as the nighttime images. However, these methods require a complex training process and multiple stages, with the training of the later stages relying heavily on the previous stages. In addition, although some datasets such as ACDC \cite{sakaridis2021acdc} and Dark Zurich \cite{sakaridis2019guided} provide corresponding pairs of images, their semantic information is not entirely consistent and needs extra alignment, making the process even more challenging.


Self-training (ST) offers a promising solution for the problem of limited labeled data by incorporating large-scale unlabeled data. Recently, some studies \cite{hoyer2022daformer,gao2021dsp} employed the mean teacher \cite{tarvainen2017mean} framework for ST, with UDA methods performing output-level alignment based on consistency constraints between the student and teacher models' target predictions. However, previous methods only utilize one-directional knowledge transfer – updating the teacher by the student – which fails to consider feedback from the teacher model in enhancing knowledge of the target domain.




In general, the exponential moving average (EMA) of the student network is usually responsible for maintaining the weights of the teacher, which are utilized for making pseudo-label predictions. This process, depicted in Fig. \ref{intro} (d), is based on the assumption that a single teacher is able to supervise the student effectively. However, this approach is much like having an elementary school teacher supervise a graduate student, because a single-teacher model is often unable to capture the significant disparities across different domains. As existing ST methods are known to have confirmation bias problems \cite{chen2022debiased}, it is likely that noise of supervised signals resulting from incorrect pseudo-labeling can accumulate during the training. For example, in Fig \ref{intro} (b), the boundary of the sidewalk is recognized as road, and the building is recognized as wall. This may be due to noisy supervision caused by the large domain gap.



Based on the above facts, we propose a new self-training viewpoint on domain adaptation that decouples multiple factors of cross-domain shift when adapting from the source domain to the target domain, and then performs circular iterations between the student and teachers. Hence, we propose a novel one-stage Dual-Teacher Bidirectional Self-training (DTBS) framework to explicitly explore the main aspects of domain shift and average dual-teacher feedback for domain adaptation. Fig. \ref{intro} (e) illustrates the conceptual diagram of our DTBS.

The DTBS framework consists of two key components: Gradual Domain Mixing (GDM) and Teachers-Student Feedback (TSF). Inspired by CuDA-Net \cite{ma2022both}, we assume that the total domain shift can be attributed to style and illumination. The style shift typically stems from the differences in hue and street style across domains. The illumination shift refers to environmental light changes from daytime to nighttime. Towards this end, GDM generates the mask used to derive the source domain patch, which is sequentially pasted onto the target domain, i.e. daytime and nighttime, while GT provides partial effective supervision of the mixed images. In addition, we add TSF to iterate a more robust student. Unlike the conventional single-teacher ST, which employs a teacher (i.e. updated with the student at moments $1,2,\dots,t-1$) to instruct the current student, our approach presents a new training manner by urging the t-moment student to smoothly learn from two teachers with complementary knowledge. Inspired by \cite{tarvainen2017mean}, we propose averaging teacher model weights instead of predictions. Ultimately, both teachers impart knowledge (style and illumination) to the student via Re-weighted EMA, thereby fostering a closed-loop system encompassing superior anti-interference capabilities. After DTBS, network can better exploit style and illumination factors and successfully segment confusing areas, such as sidewalk and building in Fig \ref{intro} (c).




The main contributions of this paper are summarized as follows:
\begin{itemize}
    \item For nighttime semantic segmentation, we propose a novel Dual-Teacher Bi-directional Self-training UDA framework, DTBS, which requires neither a image translation network nor multiple training stages.
    \item To address the issue of domain shift in style and illumination, we propose Gradual Domain Mixing (GDM) for smooth knowledge transfer. To alleviate the confirmation bias issue in conventional self-training, we propose Teachers-Student Feedback (TSF) to integrate complementary knowledge of teachers from different domains to iterate student. 
    \item Our proposed method shows excellent performance on two challenging UDA benchmarks, ACDC night and Dark Zurich, achieving a 5.0\% and 1.2\% mIou improvement relative to the DAFormer.
\end{itemize}

\section{Related Work}

Our work is an extension of UDA methods, which are mainly two kinds of adversarial learning and self-training. In this paper, we improve the conventional self-training, so we briefly review some related work in self-training methods.

\subsection{Self-training}
Self-training is a widely used semi/unsupervised learning strategy that generates pseudo-labels of unlabeled data. It has two main categories: online self-training and offline self-training. 

\textbf{Online self-training} mainly adds extra training in each iteration. Nikita $et \: al.$ \cite{araslanov2021self} trained the model with data augmented by photometric noise, flipping, and scaling. DACS \cite{tranheden2021dacs} used a class mix for data augmentation. 

\textbf{Offline self-training} utilizes the output pseudo-labels from the fully trained model to update other models and repeats this process several times. CBST \cite{zou2018unsupervised} proposed a new UDA framework based on an iterative self-training that can be solved by alternatively generating pseudo-labels on the target data and retraining the model with these labels. CRST \cite{zou2019confidence} treated pseudo-labeling as a continuous potential variable jointly optimized by alternating optimization.

\subsection{Intermediate Domain-based Methods}
 Introducing an intermediate domain is a general approach for domain adaptation based on data augmentation, i.e. by additionally training images at a semantic level between the source and target domains. Its primary types include real intermediate domain, rendering synthetic images and domain mixing.

\textbf{Real intermediate domain} includes Dark Zurich's day and twilight, and reference images of the ACDC dataset, both are based on real scenes corresponding to the target domain collected by GPS localization. DMAda \cite{dai2018dark} is the first semantic model trained to adapt gradually to nighttime scenes using the twilight moment as a bridge. GCMA \cite{sakaridis2019semantic} and MGCDA \cite{sakaridis2020map} are based on curriculum learning frameworks using inter-temporal pairs of real images to guide the inference.


\textbf{Rendering synthetic images} artificially creates intermediate domain images with source and target domain styles simultaneously, such as CycleGAN \cite{zhu2017unpaired}, Color Transfer \cite{reinhard2001color} and FDA \cite{yang2020fda}. Thus some works like DANNet \cite{wu2021dannet}, CDAda \cite{xu2021cdada}, and CCDistill \cite{gao2022cross} embedded rendering synthetic images methods in UDA to achieve smooth semantic knowledge transfer from day to night.



\textbf{Domain mixing} has been well studied to improve the robustness of models in other tasks, e.g. image classification \cite{yun2019cutmix} and object detection \cite{olsson2021classmix}.  Researchers \cite{tranheden2021dacs,zhou2022context}  studied cross-domain mixup in UDA. DACS \cite{tranheden2021dacs} proposed a method to mix source and target samples by ClassMix. CAMix \cite{zhou2022context} proposed to exploit the vital clue of context-dependency as prior knowledge to enhance the adaptability toward the target domain.


\begin{figure*}[htbp]
    
	\small
	\setlength{\abovecaptionskip}{0.1cm}
	\centering
	\includegraphics[width=18cm]{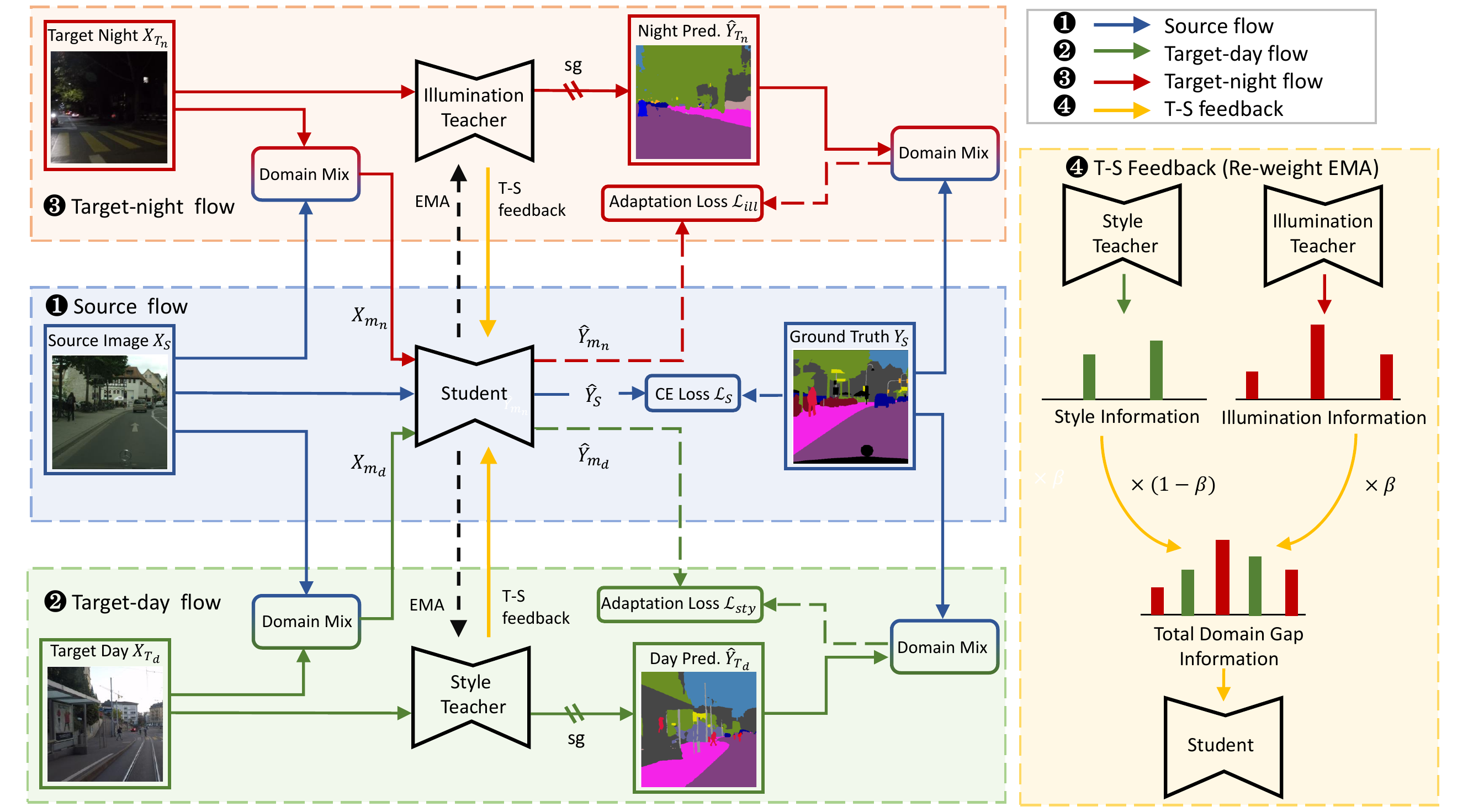}
	\caption{Overview of the proposed Dual-teacher Bidirectional Self-training (DTBS) architecture. Source flow \ding{182}$\rightarrow$ Target-day flow \ding{183}$\rightarrow$ Target-night flow \ding{184}$\rightarrow$ T-S Feedback \ding{185} are the four sub-flows that make up each iteration, with $sg$ standing for no gradient backward propagation. The first three workflows achieve smooth domain adaptation. T-S feedback integrates knowledge to iteratively refine the student.} 
        \label{overview}
\end{figure*}

\subsection{Mean Teacher-based Methods}

For single teacher's setups, SEANET \cite{xu2019self} first attempted to introduce a self-integrating model into domain adaptation for semantic segmentation based on the mean teacher, and Zhou et al. \cite{zhou2022uncertainty} presented an uncertainty-guided consistency loss with a dynamic weighting scheme based on the potential uncertainty information of the target sample. 


Although these works performe well, they all update the teacher network weights based on the student network, ignoring the fact that the teacher's learned knowledge can benefit the student network, especially when multiple teacher models decouple domain differences.

For multiple teachers' setups, DDB \cite{chen2022deliberated} and MetaTeacher \cite{wang2022metateacher} only ensemble the outputs of two teachers as stronger guidance to supervise the student, while our DTBS update the two teachers model PARAMETERS to the student by EMA, which can pass different kinds of knowledge to the student more effectively.



\section{Methods}

\subsection{Overview and Notations}

We follow the UDA protocol \cite{zou2018unsupervised, vu2019advent} and set the source domain images $X_S \in S$ and corresponding labels $Y_S$. And the target domain $T$ includes unlabeled daytime ($X_{T_d}$) and nighttime ($X_{T_n}$) images. The $X_{T_d}$ and $X_{T_n}$ were taken at different times in GPS-matched locations, resulting in significant differences in illumination, but consistent semantic information. Although both $X_S$ and $X_{T_d}$ are daytime images, significant differences in urban styles and image tones constitute the total style difference. Thus, we assume that the domain discrepancy includes both style and illumination. 


Online \cite{hoyer2022daformer} or offline \cite{zou2018unsupervised,yang2020fda} pseudo-labels can be produced. We choose the online ST for its properties of single stage training and simple operation. Some methods \cite{hoyer2022daformer} naively repurposed the mean teacher \cite{tarvainen2017mean} framework, just considering the knowledge transfer from the student to the teacher, i.e. updating teacher $h_\phi$ based on student $g_\theta$. Thus, the teacher model may accrue negative transfer. To alleviate this deficiency, we propose a bidirectional knowledge transfer architecture that decomposes the style-grid-illumination domain shift, aiming to enhance performance in UDA tasks .

Fig. \ref{overview} shows the overview of our proposed architecture, where the specific processes are elaborated in Sec. \ref{total}. We generate mask $M$ by the GDM strategy (proposed in Sec. \ref{section B}), thereby decoupling the two factors affecting the domain shift --- style and illumination, which mixes $X_S$ and $X_{T_d}$  to generate daytime mixed images $X_{m_d}$, and mixes $X_S$ and $X_{T_n}$ to generate nighttime mixed images $X_{m_n}$. Then $X_{T_d}$ and $X_{T_n}$ are fed to the style teacher model ($h_{\phi^S}$) and the illumination teacher model ($h_{\phi^I}$), sequentially. The predicted segmentation maps $\hat{Y}_{T_d}$ and $\hat{Y}_{T_n}$ yield pseudo labels (described in Sec. \ref{section C}). Notably, the weights of the teachers $h_{\phi^S}$ and $h_{\phi^I}$ are the EMA of the weights of the student $g_\theta$. Since our final test is nighttime, each train iteration ends with TSF strategy (introduced in Sec. \ref{section D}), where the illumination teacher gives more feedback than the style teacher.

Regarding other notations, $\phi^S_t,\phi^I_t,\theta_t$ are the weights of style teacher, illumination teacher and student model at the $t$-th iteration, respectively. $(h,w)$ denotes the pixel with height $h$ and width $w$, $H$ and $W$ are the height and width of the image.

\subsection{Gradual Domain Mixing}
\label{section B}

\subsubsection{Algorithm for Strategy}
\begin{figure}[h]
    
	\small
	\setlength{\abovecaptionskip}{0.1cm}
	\centering
	\includegraphics[width=7cm]{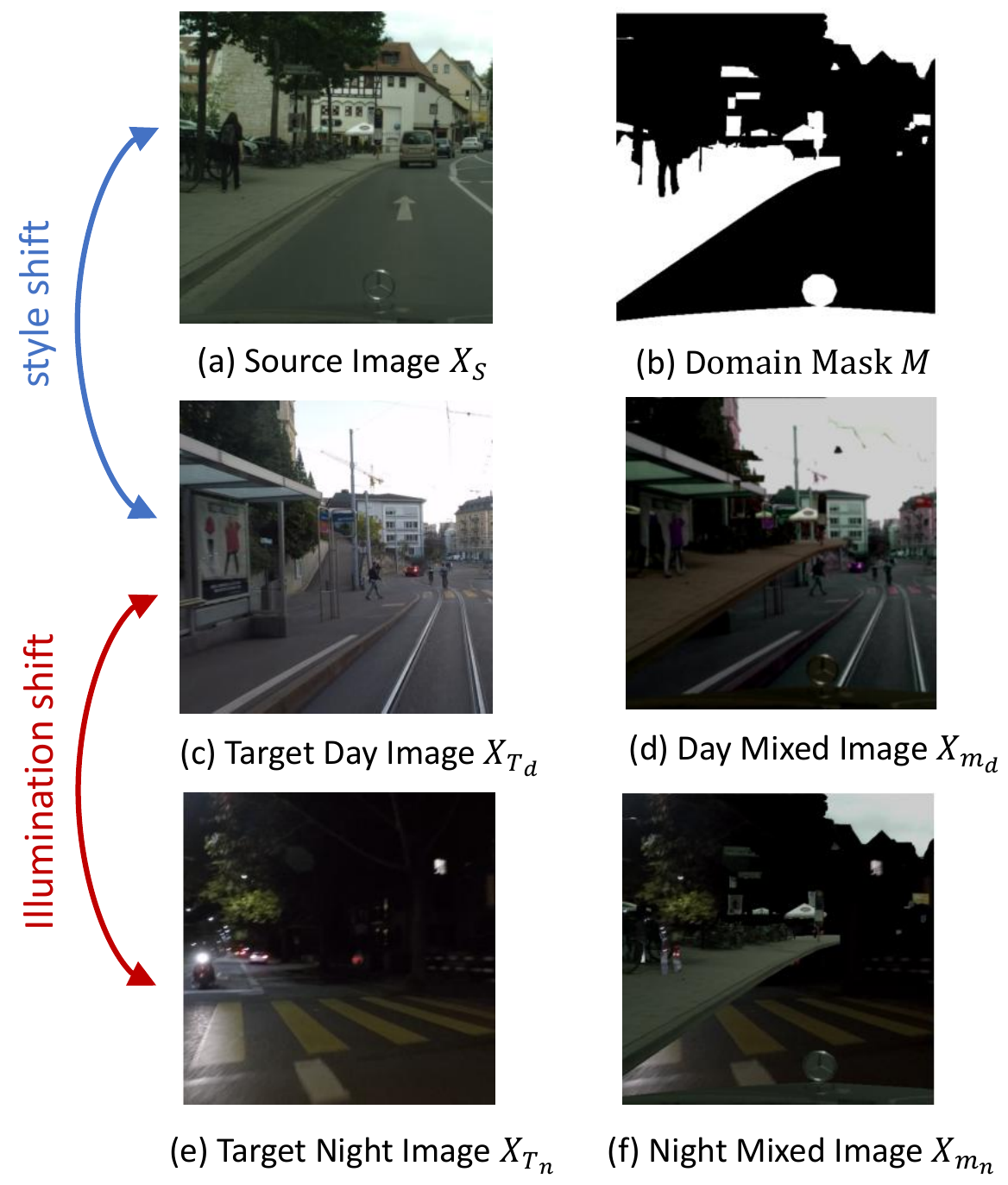}
	\caption{Our goal is to transfer the knowledge from a labeled image (a) to an unlabeled image (e). Considering that direct knowledge transfer is challenging due to the large domain shift, by introducing an intermediate domain daytime image (c), we decompose the overall domain shift into style and illumination. To further smooth the knowledge transfer, the source domain image (a) is mixed with the target domain daytime (c) and nighttime (e) image to produce two mixed images (d) (f).} 
        \label{221}
\end{figure}

It makes sense that adapting the style variance is simpler than adapting the light variance. With this in mind, we the propose Gradual Domain Mixing (GDM) to minimize the domain gap between the source domain and target night domain step by step. In each iteration, we first select a random source image $X_S$ from $S$ through RCS \cite{hoyer2022daformer} strategy, and then choose half of its classes and corresponding image blocks as the candidate patch for subsequent pasting. In this way, we can ensure that the candidate patch has both frequent classes and long-tail classes. Then we paste the candidate patch on the daytime and nighttime images in the target domain. And we define $ M \in\{0,1\}^{H \times W}$ as a binary mask in which $M(h, w)=1$ when the pixel position $(h, w)$ of $X_S$ belongs to the selected classes, and $M(h, w)=0$ otherwise. Specifically, given the source image $X_S$, the target image $X_{T_d},X_{T_n}$ and the corresponding binary mask $M$, the mixed source image $X_{m_d}$ and $X_{m_n}$ can be obtained by:
\begin{equation}
\begin{aligned}
& X_{m_d}=(1-M) \odot X_{T_d}+M \odot X_S \\
& X_{m_n}=(1-M) \odot X_{T_n}+M \odot X_S
\end{aligned}
\end{equation}
where $\odot$ denotes element-by-element multiplication. In addition, we show the visualization example in Fig. \ref{221}. Unlike previous work in \cite{wu2021one} where target domain coarsely aligned day-night image pairs are required to be sampled each time, we randomly sampled $X_{T_d}$ and $X_{T_n}$ from target domain's day and night datasets, avoiding the cumbersome alignment process while achieving good performance.

Due to the inconsistent stylistic distributions in the mixed images produced by ClassMix, the performance of the adaptation may suffer. To learn more robust features, we follow DACS \cite{tranheden2021dacs} and use color jittering and Gaussian blur as data enhancements.

\begin{table*}[ht]
\caption{Comparison to the state of the art in Cityscapes$\rightarrow$ACDC night domain adaptation on the ACDC test (night).}
 \setlength{\tabcolsep}{2mm}{
\resizebox{\textwidth}{!}{

\begin{tabular}{lcccccccccccccccccccc}
\hline
                   & \rot{road} &\rot{sidewalk} & \rot{building} & \rot{wall} & \rot{fence} & \rot{pole} & \rot{traffic light} & \rot{traffic sign} & \rot{vegetation} & \rot{terrain} & \rot{sky}  & \rot{person} & \rot{rider} & \rot{car}  & \rot{truck} & \rot{bus}  & \rot{train} & \rot{motorcycle} & \rot{bicycle} & \rot{Mean} \\ \hline
AdaptSegNet \cite{lin2019adapting}        & 84.9 & 39.9     & 66.8     & 17.2 & 17.7  & 13.4 & 17.6          & 16.4         & 39.6       & 16.1    & 5.7  & 42.8   & 21.4  & 44.8 & 11.9  & 13.0 & 39.1  & 27.5       & 28.4    & 29.7 \\
ADVENT \cite{vu2019advent}             & 86.5 & 45.3     & 60.8     & 23.2 & 12.5  & 15.4 & 18.0          & 19.4         & 41.2       & 18.3    & 2.7  & 43.8   & 21.3  & 61.6 & 12.6  & 19.1 & 43.0  & 30.2       & 27.6    & 31.7 \\
CMAda \cite{sakaridis2018model}              & 82.6 & 25.4     & 53.9     & 10.1 & 11.2  & 30.5 & 36.7          & 30.0         & 38.7       & 16.5    & 0.1  & 46.0   & 26.2  & 65.8 & 13.9  & 50.9 & 20.4  & 24.8       & 23.8    & 32.0 \\
DMAda \cite{dai2018dark}              & 74.7 & 29.5     & 49.4     & 17.1 & 12.6  & 31.0 & 38.2          & 30.0         & 48.0       & 22.8    & 0.2  & 47.0   & 25.4  & 63.8 & 12.8  & 46.1 & 23.1  & 24.7       & 24.6    & 32.7 \\
BDL \cite{li2019bidirectional}                & 87.1 & 49.6     & 68.8     & 20.3 & 17.5  & 16.7 & 19.9          & 24.1         & 39.1       & 23.7    & 0.2  & 42.0   & 20.4  & 63.7 & 18.0  & 27.0 & 45.6  & 27.8       & 31.3    & 33.8 \\
MGCDA \cite{sakaridis2020map}              & 74.5 & 52.5     & 69.4     & 7.7  & 10.8  & 38.4 & 40.2          & 43.3         & 61.5       & 36.3    & 37.6 & 55.3   & 25.6  & 71.2 & 10.9  & 46.4 & 32.6  & 27.3       & 33.8    & 40.8 \\
GCMA \cite{sakaridis2019guided}               & 78.6 & 45.9     & 58.5     & 17.7 & 18.6  & 37.5 & \textbf{43.6}          & 43.5         & 58.7       & 39.2    & 22.5 & 57.9   & 29.9  & 72.1 & 21.5  & 56.3 & 41.8  & 35.7       & 35.4    & 42.9 \\
DANNet (PSPNet) \cite{wu2021dannet}     & 90.7 & \textbf{78.7}    & 75.6     & 35.9 & \textbf{28.8}  & 26.6 & 31.4          & 30.6         & \textbf{70.8}       & 39.4    & \textbf{78.7} & 49.9   & 28.8  & 65.9 & 24.7  & 44.1 & 61.1  & 25.9       & 34.5    & 47.6 \\
DAFormer \cite{hoyer2022daformer} & 89.4 & 55.2     & 69.5     & 36.1 & 12.9  & 44.2 & 28.0          & 42.7         & 56.9       & 40.5    & 52.8 & \textbf{59.0}   & 29.9  & 73.2 & 22.0  & 43.7 & 82.4  & 41.9       & \textbf{47.4}    & 48.8 \\ \hline

DTBS         & 92.4 & 64.4     & \textbf{76.3}     & \textbf{37.5} & 25.7  & 45.9 & 18.1          & 43.1         & 57.9       & 40.1    & 59.7 & 58.1   & 31.3  & \textbf{77.6} & \textbf{43.2}  & \textbf{73.3} & \textbf{86.2}  & 44.2       & 46.5    & \textbf{53.8} \\ 

DTBS (with TSF-E)  & \textbf{92.8} & 66.1     & 75.6     & \textbf{37.5} & 27.4  & \textbf{46.4} & 17.6          & \textbf{44.3}         & 60.0       & \textbf{40.8}    & 60.7 & 58.3   & \textbf{31.4}  & 77.5 & 38.6  & 62.6 & 83.9  & \textbf{44.6}       & 46.2    & 53.3 \\ \hline

\end{tabular}}}
\label{com_acdc}
\end{table*}

\subsection{Teachers-Student Feedback Domain Adaptation}
\label{section C}
\subsubsection{Output Level Alignment}
The source domain images $X_S,$ mixed daytime images $X_{m_d}$ and mixed nighttime images $X_{m_n}$ are fed to the student network $g_\theta$, while the original target domain images $X_{T_d},X_{T_n}$ are fed to the style teacher model $h_{\phi^S}$ and illumination teacher model $h_{\phi^I}$, respectively. Of importance is that we first perform target-day flow, followed by target-night flow. The parameters $\theta$ of the student network are updated by gradient backpropagation, while the parameters $\phi_d, \phi_n$ of the two teacher networks are updated in their corresponding flow by the EMA to obtain stable pseudo labels:

\begin{equation}
\begin{aligned}
& \phi^S_{t} \leftarrow \alpha \phi^S_{t-1}+(1-\alpha) \theta_{t-1} \\
& \phi^I_{t} \leftarrow \alpha \phi^I_{t-1}+(1-\alpha) \theta_{t-1}
\end{aligned}
\end{equation}
where $\alpha$ denotes the EMA decay coefficient and $t$ denotes train iterations.
After obtaining the segmentation prediction map of the source domain image $X_S$, the student network $g_\theta$ is trained with the categorical cross-entropy (CE) loss.

\begin{equation}
    \mathcal{L}_S^{}=-\sum_{j=1}^{H \times W} \sum_{c=1}^C Y_S^{(j, c)} \log g_\theta\left(X_S^{}\right)^{(j, c)}
\end{equation}
where $C$ represents the number of classes, and $Y_S$ denotes the GT semantic labels of $X_S$.
To reduce the gap between domains' style distributions, pseudo-labels and their quality estimates $q_{T_d}$ \cite{hoyer2022daformer} were used to additionally train the student $g_\theta$ on the target domain day. After obtaining the predicted semantic map $\hat{Y}_{m_d}$ of $X_{m_d}$ from $g_\theta$, we use a weighted cross-entropy loss to train the network.

\begin{equation}
\begin{gathered}
\mathcal{L}_{sty}^{}=-\sum_{j=1}^{H \times W} \sum_{c=1}^C Y_S^{(j, c)} \log \left(\hat{Y}_{m_d}\right)^{(j, c)} \odot M-\\
\sum_{j=1}^{H \times W} \sum_{c=1}^C q_{T_d}^{} \hat{Y}_{T_d}^{(j, c)} \log \left(\hat{Y}_{m_d}\right)^{(j, c)} \odot(1-M) 
\end{gathered}
\end{equation}
where $q_{T_d}$ denotes the ratio of pixels for which $Y_{T_d}$ exceeds the maximum softmax probability threshold $\tau$. The $q_{T_d}$ and $\tau$ are defined in the method proposed by Lukas et al. \cite{tranheden2021dacs}. Once the model has minimized the style shift, it will be easier to adapt to the illumination shift.
 Thus, the same workflow additionally trains the student in target domain night. With the predicted semantic map $\hat{Y}_{m_n}$ of $X_{m_n}$ from $g_\theta$, we also use a weighted cross-entropy loss to train the network.

\begin{equation}
\begin{gathered}
\mathcal{L}_{ill}^{}=-\sum_{j=1}^{H \times W} \sum_{c=1}^C Y_S^{(j, c)} \log \left(\hat{Y}_{m_n}\right)^{(j, c)} \odot M-\\
\sum_{j=1}^{H \times W} \sum_{c=1}^C q_{T_n}^{} \hat{Y}_{T_n}^{(j, c)} \log \left(\hat{Y}_{m_n}\right)^{(j, c)} \odot (1-M)
\end{gathered}
\end{equation}
where $\hat{Y}_{T_d}$ and $\hat{Y}_{T_n}$ denote the predicted segmentation maps of the style and illumination teacher network output, respectively.

\begin{table*}[ht]
\caption{Comparison to the state of the art in Cityscape$\rightarrow$Dark Zurich domain adaptation on the Dark-Zurich-test.}
 \setlength{\tabcolsep}{2mm}{
\resizebox{\textwidth}{!}{
\begin{tabular}{lcccccccccccccccccccc}
\hline

                   & \rot{road} &\rot{sidewalk} & \rot{building} & \rot{wall} & \rot{fence} & \rot{pole} & \rot{traffic light} & \rot{traffic sign} & \rot{vegetation} & \rot{terrain} & \rot{sky}  & \rot{person} & \rot{rider} & \rot{car}  & \rot{truck} & \rot{bus}  & \rot{train} & \rot{motorcycle} & \rot{bicycle} & \rot{Mean} \\ \hline
DeepLabv2 \cite{chen2017deeplab}             & 79.0 & 21.8     & 53.0     & 13.3 & 11.2  & 22.5 & 20.2          & 22.1         & 43.5       & 10.4    & 18.0 & 37.4   & 33.8  & 64.1 & 6.4   & 0.0  & 52.3  & 30.4       & 7.4     & 28.8 \\
ADVENT \cite{vu2019advent}               & 85.8 & 37.9     & 55.5     & 27.7 & 14.5  & 23.1 & 14.0          & 21.1         & 32.1       & 8.7     & 2.0  & 39.9   & 16.6  & 64.0 & 13.8  & 0.0  & 58.8  & 28.5       & 20.7    & 29.7 \\
AdaptSegNet \cite{lin2019adapting}          & 86.1 & 44.2     & 55.1     & 22.2 & 4.8   & 21.1 & 5.6           & 16.7         & 37.2       & 8.4     & 1.2  & 35.9   & 26.7  & 68.2 & 45.1  & 0.0  & 50.1  & 33.9       & 15.6    & 30.4 \\
BDL \cite{li2019bidirectional}                  & 85.3 & 41.1     & 61.9     & 32.7 & 17.4  & 20.6 & 11.4          & 21.3         & 29.4       & 8.9     & 1.1  & 37.4   & 22.1  & 63.2 & 28.2  & 0.0  & 47.7  & 39.4       & 15.7    & 30.8 \\
DANNet (DeepLabv2) \cite{wu2021dannet}   & 88.6 & 53.4     & 69.8     & 34.0 & 20.0  & 25.0 & 31.5          & 35.9         & 69.5       & 32.2    & 82.3 & 44.2   & 43.7  & 54.1 & 22.0  & 0.1  & 40.9  & 36.0       & 24.1    & 42.5 \\
DANIA (DeepLabv2) \cite{wu2021one}    & 89.4 & 60.6     & 72.3     & 34.5 & 23.7  & 37.3 & 32.8          & 40.0         & \textbf{72.1}       & 33.0    & \textbf{84.1} & 44.7   & 48.9  & 59.0 & 9.8   & 0.1  & 40.1  & 38.4       & 30.5    & 44.8 \\


DMAda (RefineNet) \cite{dai2018dark}      & 75.5 & 29.1     & 48.6     & 21.3 & 14.3  & 34.3 & 36.8          & 29.9         & 49.4       & 13.8    & 0.4  & 43.3   & 50.2  & 69.4 & 18.4  & 0.0  & 27.6  & 34.9       & 11.9    & 32.1 \\
GCMA (RefineNet) \cite{sakaridis2019guided}     & 81.7 & 46.9     & 58.8     & 22.0 & 20.0  & 41.2 & \textbf{40.5}          & \textbf{41.6}         & 64.8       & 31.0    & 32.1 & 53.5   & 47.5  & 75.5 & 39.2  & 0.0  & 49.6  & 30.7       & 21.0    & 42.0 \\
MGCDA (RefineNet) \cite{sakaridis2020map}    & 80.3 & 49.3     & 66.2     & 7.8  & 11.0  & 41.4 & 38.9          & 39.0         & 64.1       & 18.0    & 55.8 & 52.1   & 53.5  & 74.7 & 66.0  & 0.0  & 37.5  & 29.1       & 22.7    & 42.5 \\
CDAda (RefineNet) \cite{xu2021cdada}    & 90.5 & 60.6     & 67.9     & 37.0 & 19.3  & 42.9 & 36.4          & 35.3         & 66.9       & 24.4    & 79.8 & 45.4   & 42.9  & 70.8 & 51.7  & 0.0  & 29.7  & 27.7       & 26.2    & 45.0 \\
DANNet (PSPNet) \cite{wu2021dannet}      & 90.4 & 60.1     & 71.0     & 33.6 & 22.9  & 30.6 & 34.3          & 33.7         & 70.5       & 31.8    & 80.2 & 45.7   & 41.6  & 67.4 & 16.8  & 0.0  & 73.0  & 31.6       & 22.9    & 45.2 \\
CCDistill (RefineNet) \cite{gao2022cross} & 89.6 & 58.1     & 70.6     & 36.6 & 22.5  & 33.0 & 27.0          & 30.5         & 68.3       & 33.0    & 80.9 & 42.3   & 40.1  & 69.4 & 58.1  & 0.1  & 72.6  & \textbf{47.7}       & 21.3    & 47.5 \\
DANIA (PSPNet) \cite{wu2021one}       & 91.5 & 62.7     & 73.9     & \textbf{39.9} & \textbf{25.7}  & 36.5 & 35.7        & 36.2         & 71.4       & \textbf{35.3}    & 82.2 & 48.0   & 44.9  & 73.7 & 11.3  & 0.1  & 64.3  & 36.7       & 22.7    & 47.0 \\


DAFormer \cite{hoyer2022daformer} & 92.7 & \textbf{64.5}     & 73.5     & 33.6 & 16.6  & \textbf{52.0} & 30.2          & 36.4         & 68.2       & 31.4   & 79.3  & 54.4 & 51.8  & \textbf{78.4 } & 69.3  & 9.7  & 89.6  & 45.2    & 37.5     & 53.4  \\ \hline

DTBS & 92.2 & 60.3     & \textbf{77.1}     & 39.2 & 18.4  & 51.4 & 30.5          & 39.6         & 66.2       & 27.5    & 79.1 & \textbf{56.4}   & \textbf{54.0}  & 76.5 & 81.1  & 9.2 & \textbf{90.4}  & 45.7       & 39.8    & 54.5 \\

DTBS (with TSF-E) & \textbf{93.1} & 62.8     & 76.9     & 39.5 & 17.9  & 51.9 & 29.0          & 40.0         & 63.9       & 27.4    & 77.7 & 56.0   & 53.6  & 78.0 & \textbf{81.3}  & \textbf{13.9} & 90.1  & 44.8       & \textbf{40.2}    & \textbf{54.6} \\ \hline

\end{tabular}}}
\label{com_dark}
\end{table*}

\subsubsection{Feedback from Teacher Models}
\label{section D}


Inspired by recent research \cite{wortsman2022model}, the ensemble of different model weights often exhibits excellent performance, we propose to construct the Re-weight EMA (eq. \ref{feed}) to let the weights of the two teacher models jointly update the student. Unlike offline distillation of knowledge from teacher models \cite{gao2022cross}, our approach can integrate information from different domains online, which requires only one stage and conserves computing resources. 

\begin{equation}
    \theta_{t+1} \leftarrow \alpha \theta_{t}+(1-\alpha) \left[\beta\phi^I_{t}+(1-\beta)\phi^S_{t}\right]\label{feed}
\end{equation}

$\beta \in(0,1)$ denotes the feedback coefficient hyperparameter, and a more significant model feedback coefficient indicates more transferred knowledge. In general, the predictive entropy of the output of the teacher model for the daytime batch compared to the nighttime batch is usually smaller, indicating a higher confidence level. Thus, the two teacher models should not provide equal feedback to the student model. Since we mainly study the nighttime scenes and illumination teacher model feedback has higher priority, we set $\beta>0.5$ here.

Furthermore, a fixed feedback coefficient only roughly averages the weights of two teachers. Considering that the prediction entropy of each teacher changes dynamically in each iteration, to achieve better EMA ensembling effects, we suggest not using prediction entropy to constrain the model, but using the ratio of the total sum of pixel-level normalized entropy of the two teachers' prediction maps to set the feedback coefficient. Therefore, an improved version of TSF, i.e. TSF-E is proposed as follows.

Given a target input image, the sum of all pixel-wise normalized entropies is defined as follows:
\begin{equation}
\begin{aligned}
&\boldsymbol{E}_{T_d}=\sum_{j=1}^{H \times W} \frac{-1}{\log (C)} \sum_{c=1}^C \boldsymbol{P}_{T_d}^{(j, c)} \log \boldsymbol{P}_{T_d}^{(j, c)}\\
&\boldsymbol{E}_{T_n}=\sum_{j=1}^{H \times W} \frac{-1}{\log (C)} \sum_{c=1}^C \boldsymbol{P}_{T_n}^{(j, c)} \log \boldsymbol{P}_{T_n}^{(j, c)}
\end{aligned}
\end{equation}
where $\boldsymbol{P}_{{T_d}}^{(h, w, c)},\boldsymbol{P}_{{T_n}}^{(h, w, c)}$ represent the soft-segmentation map obtained through softmax layer output of $X_{T_d}$ and $X_{T_n}$. Based on our previous discussion on teacher knowledge feedback, we redefine the feedback coefficient as follows:
\begin{equation}
\beta=\frac{\boldsymbol{E}_{T_n}}{\boldsymbol{E}_{T_n}+\boldsymbol{E}_{T_d}}
\end{equation}

Therefore, student can dynamically receive integrated teacher knowledge to avoid the issue of a teacher passing on too much incorrect knowledge to the student.

\begin{figure*}[h]
 
	\small
	\centering
	\includegraphics[width=\textwidth]{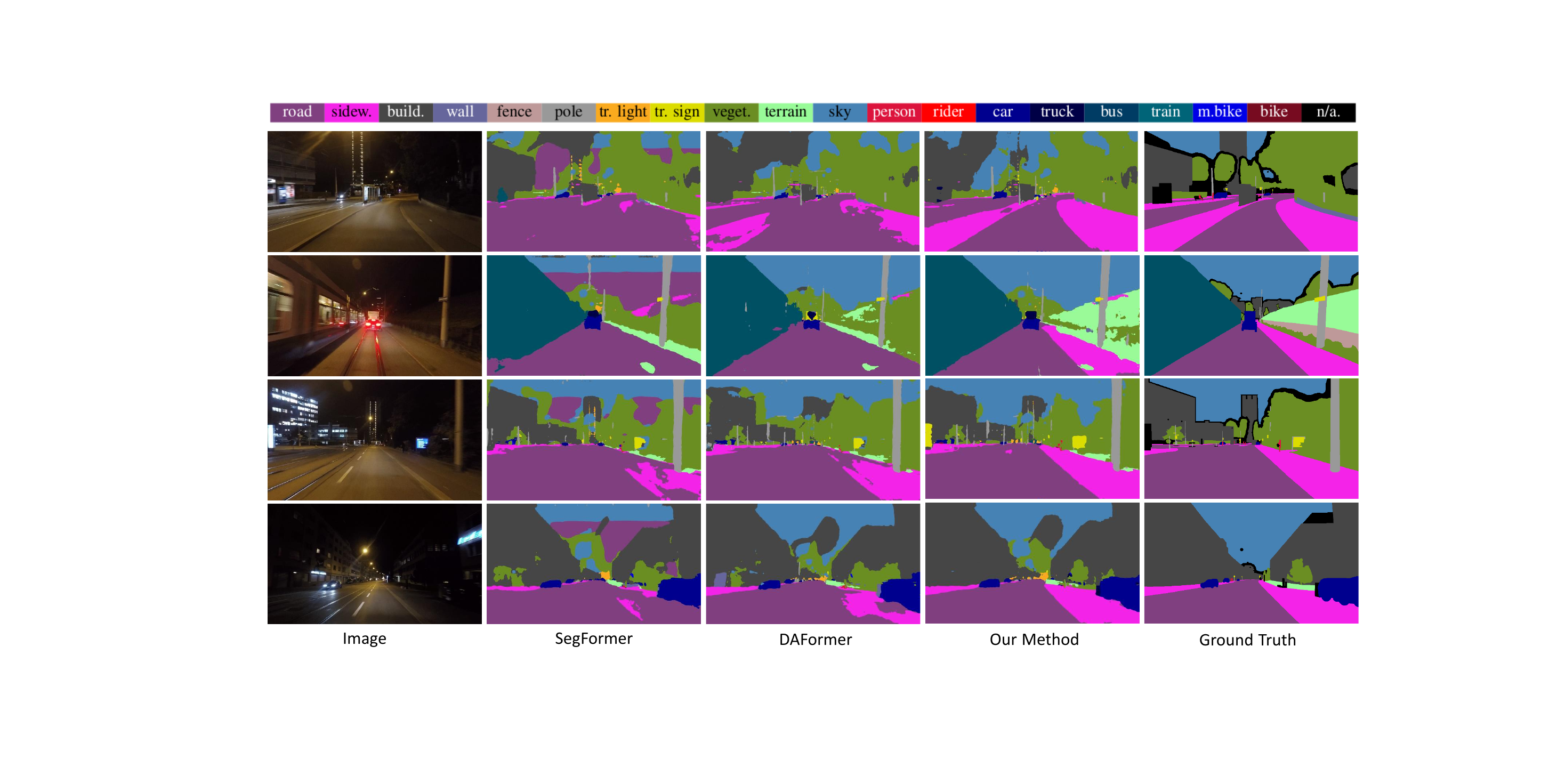}
	\caption{Some visual segmentation results of ACDC night val for the Cityscapes$\rightarrow$ACDC night task. With the teacher model parameters feedback, our method is superior at predicting the street-side structures (building, fence) and easily-confused classes (traffic sign, terrain).} 
        \label{com_pics}

\end{figure*}

\subsection{Training Workflows}
\label{total}

\textbf{Total loss.}  As our motif shows, the style and illumination shifts cumulatively constitute the overall shift. We use $\mathcal{L}_{sty}$ to denote the style shift between the domain $S$ and $T_d$. Since the illumination shift between the domain $S$ and $T_n$ is significantly larger than the style shift and we performed adaptation to style shift, we reasonably assume that we take $\mathcal{L}_{ill}$ to denote the illumination shift between the source and target nighttime domain. Therefore, we design the total loss function as follows.

\begin{equation}
\mathcal{L}_{\text {total }}=\mathcal{L}_S+\mathcal{L}_{sty}+\mathcal{L}_{ill}\label{total loss}
\end{equation}

\textbf{Training pipeline.}  Fig. \ref{overview} depicts the whole training process; the four sub-workflows are used in each iteration. In the first workflow, supervised training is performed on the student model in the source domain. In the second workflow, the style teacher model instructs student to learn style features. In the third workflow, the style teacher model is frozen, and the illumination teacher model instructs the student to learn the illumination features. The loss function of the first three workflows training can be centrally represented as eq. \ref{total loss}. Finally, the complementary knowledge of the two teachers jointly iterates the students, as shown in eq. \ref{feed}. Teachers-Student Feedback uses both the EMAs of the two teachers’ weights for iterating a student network with robustness.

\section{Experiments}
\subsection{Implementation Details} 
\subsubsection{Datasets} 

Cityscapes \cite{cordts2016cityscapes} is used for the street scenes, with 2,975 images for training, 500 images for validation, and 1,525 images for test. There are 19 categories of pixel-level annotations, and both the original images and annotations have a resolution of 2,048 × 1,024 pixels.

Dark Zurich \cite{sakaridis2019guided} is a street scene taken in Zurich with 3,041 daytime, 2,920 twilight, and 2,416 nighttime images, all of which are unlabeled images with 1,920 × 1,080 resolution. Dark Zurich also contains 201 manually annotated night images, of which 151 (Dark Zurich-test) are used for test, and 50 (Dark Zurich-val) are used for validation. 

ACDC \cite{sakaridis2021acdc} contains a total of 4,006 images for four adverse conditions (fog, rain, nighttime and snow). The nighttime has pixel-level annotations with 400 train images, 106 validation images, and 500 test images. It has a similar style and appearance to Dark Zurich. Therefore, we use ACDC-night-test to evaluate our network on domain adaptation further.

\subsubsection{Experimental Settings}

We implement our proposed method with PyTorch on a single Nvidia 3090 GPU. Our technique is based on the mmsegmentation \cite{contributors2020mmsegmentation} framework. We choose DAFormer \cite{hoyer2022daformer} as our baseline semantic segmentation network, which was pre-trained on the ImageNet dataset with SegFormer \cite{xie2021segformer} as the backbone. And DAFormer was trained with a batch of 512 × 512 random crop for 40k iterations. We trained network with AdamW \cite{loshchilov2017decoupled}, a learning rate of $6 \times 10^{-5}$ for the encoder and $6 \times 10^{-4}$ for the decoder. Following the settings of DACS \cite{tranheden2021dacs}, we used the same data enhancement parameters and set $\alpha= 0.99$ and $\tau = 0.968$. The RCS temperature was set to $T = 0.01$ to maximize the sampled pixels for the class with rare pixels.

\begin{figure*}[h]
 
	\small
	\centering
	\includegraphics[width=16.8cm]{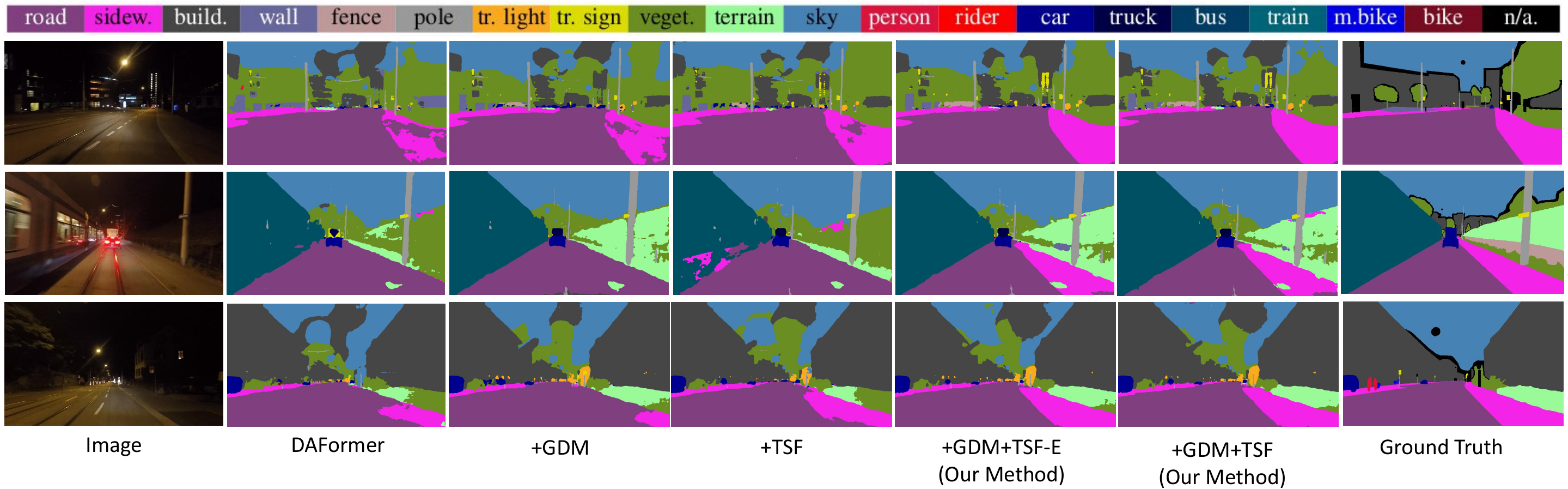}
	\caption{Qualitative components ablation experimental sample comparison of ACDC night val for Cityscapes$\rightarrow$ACDC night.} 
        \label{qua xiao}

\end{figure*}

\subsection{Comparison to the State of the Art in UDA}
\subsubsection{ACDC night}
We compare several state-of-the-art methods on the ACDC night test in Tab. \ref{com_acdc}. Adapting Cityscapes to ACDC night, our approach achieves mIoU of 53.8\%, which boosts the mIoU of baseline by 5\%. In addition, we also find improvements for some classes with similar textures (road, sidewalk, building) due to our adaptation strategies. Moreover, our approach has shown significant cross-domain generalization capability for dynamic classes. These dynamic classes include person, rider, truck, and bus. It is increasingly important in autonomous driving requirements.


We present a qualitative comparison of our method with SOTAs. Our method consistently produces more accurate segmentation maps. Four sample visualizations on the ACDC-night-val are shown in Fig. \ref{com_pics}, where we observe that DAFormer incorrectly predicts sidewalk areas as road. Based on our interpretation of the style shift, the model is more capable of generalizing to various styles of buildings due to our decoupling strategy (a 6.8\% improvement of mIoU in building). This is consistent with our hypothesis that the difference in architectural style is one of the main reasons for the style shift.

In autonomous driving, for example, person and car are dynamic, whereas building and pole are static. For static classes, TSF-E offers more significant performance improvements, while dynamic classes performance improves more when using TSF.

\subsubsection{Dark Zurich}
In Tab. \ref{com_dark}, we benchmarked our method on the Dark Zurich test. In Dark Zurich test, our method significantly outperforms DAFormer. We observe a significant improvement in the prediction accuracy of our method for the building, 3.4\% mIoU higher than baseline. In some scenarios, there are fewer people and cars, which are called ‘rare dynamic classes’. Furthermore, our method is superior in predicting some rare dynamic classes (truck, bus, bicycle by 12.0\%, 4.2\% and 2.7\% mIoU respectively).

\subsection{Ablation Study and Further Analysis}

\subsubsection{Ablation of Components}

To investigate impact of the different components of our approach, we performed an ablation study on the Cityscapes to ACDC night setup, as shown in Tab. \ref{com xiao}. Using the TSF alone brings a gain of 1.1\%. With GDM, a gain of 2.83\% mIoU can be observed. Our proposed method combines the advantages of both and two variants of TSF brings a total gain of 4.46\% and 4.95\% to the baseline model.

We also show some subjective segmentation results in Fig. \ref{qua xiao}. They indicate that segmentation results improve as more components are included in our method.

\begin{table}

  \centering
  \caption{Quantitative evaluation of component ablation on the ACDC night test.}
    \begin{tabular}{l|cc}
    \toprule
    Method & mIoU(\%) & Gain(\%) \\
    \midrule
    Baseline & 48.82 & - \\
    Baseline+TSF & 49.92 & 1.10 \\
    Baseline+GDM & 51.65 & 2.83 \\
    Baseline+GDM+TSF-E & 53.28 & 4.46 \\
    Baseline+GDM+TSF & 53.77 & 4.95 \\
    \bottomrule
    \end{tabular}%
  \label{com xiao}%
\end{table}


\subsubsection{Ablation of Feedback Coefficient $\beta$}

Tab. \ref{beta xiao} shows the sensitivity of our method to different values of its unique hyperparameter $\beta$ (feedback coefficient). When beta=0.8, the model exhibits the best performance, i.e. 40.64\% mIoU. When beta=0, the feedback is entirely contributed by the style teacher, which only obtains 39.81\% mIoU. When beta=1, the illumination teacher entirely contributes the feedback, which only obtains 39.15\% mIoU. In addition, beta variation interferes less with model performance, and the method is robust to changes in $\beta$.
\begin{table}
  \centering
    \caption{Hyper-parameter study of the feedback coefficient $\beta$.}
    \begin{tabular}{cccccc}
    \toprule
    \multicolumn{6}{c}{Cityscapes to ACDC night val} \\
    \midrule
    $\beta$  & 0     & 0.7   & 0.8   & 0.9   & 1 \\
    mIoU(\%)  & 39.81 & 40.09 & 40.64 & 39.28 & 39.15 \\
    \bottomrule
    \end{tabular}%
  \label{beta xiao}%
\end{table}

\subsubsection{Ablation of Segmentation Architectures}
We perform DTBS with original TSF and TSF-E. Tab. \ref{arcs} presents the UDA performance of different segmentation architectures on the Cityscape to ACDC night test. We have selected some typical semantic segmentation architectures based on CNN and Transformer. Our UDA method exhibits high adaptability to various architectures.

\begin{table}

  \centering
  \caption{Comparison of the mIoU (\%) on the ACDC night test. Set of segmentation architectures for DAFormer UDA method and our UDA method.}
\begin{tabular}{c|ccccc}
\hline
{\color[HTML]{000000} }                               &                          & \multicolumn{2}{c}{Our UDA} & \multicolumn{2}{c}{Our UDA (TSF-E)} \\ \cline{3-6} 
\multirow{-2}{*}{{\color[HTML]{000000} Architecture}} & \multirow{-2}{*}{Origin} & mIou          & Gain        & mIou             & Gain            \\ \hline
DeepLabV2 \cite{chen2017deeplab}                                  & 37.16                    & 37.62         & 0.46        & 38.25            & 1.09            \\
DANNet \cite{wu2021dannet}                                         & 36.61                    & 42.14         & 5.53        & 40.88            & 4.27            \\
SegFormer \cite{xie2021segformer}                                         & 45.02                    & 48.16         & 3.14        & 46.5             & 1.48            \\
DAFormer \cite{hoyer2022daformer}                                              & 48.82                    & 53.77         & 4.95        & 53.28            & 4.46            \\ \hline
\end{tabular}
\label{arcs}
\end{table}

\section{Conclusion}
We propose DTBS, an extension of the self-training-based UDA method, which achieves good performance in domain adaptation. The overall is divided into two components:(1)Adapting daytime image to nighttime image learning. (2)Teacher model knowledge ensembling feedback to student model.

To implement component 1, we propose GDM, which makes advantage of intermediate domain construction methods for mixed images and natural images complementarily. To implement component 2, we propose TSF, which allows the teacher models with different domain gap knowledge to be integrated into the student model. We apply DTBS to DAFormer and report the latest normal to adverse domain adaptation results for semantic segmentation.


\textbf{Limitations} (1) Our UDA method employs DAFormer \cite{hoyer2022daformer} as the baseline model, which can be replaced by other state-of-the-art domain adaptation methods based on Dual-Teacher self-training. (2) We focus on semantic segmentation performance in the nighttime scene and can conduct more detailed analyses on other adverse conditions to validate the correctness of teacher knowledge ensembling.

\ack This work is supported in part by the National Key R\&D Program of China (2021YFC3320301) and Joint Funds of the Zhejiang Provincial Natural Science Foundation of China (LTY22F020001).

\bibliography{ecai}

\begin{thebibliography}{10}

\bibitem{araslanov2021self}
Nikita Araslanov and Stefan Roth, `Self-supervised augmentation consistency for
  adapting semantic segmentation', in {\em Proceedings of the IEEE/CVF
  Conference on Computer Vision and Pattern Recognition}, pp. 15384--15394,
  (2021).

\bibitem{chen2022debiased}
Baixu Chen, Junguang Jiang, Ximei Wang, Jianmin Wang, and Mingsheng Long,
  `Debiased pseudo labeling in self-training', {\em arXiv preprint
  arXiv:2202.07136}, (2022).

\bibitem{chen2017deeplab}
Liang-Chieh Chen, George Papandreou, Iasonas Kokkinos, Kevin Murphy, and Alan~L
  Yuille, `Deeplab: Semantic image segmentation with deep convolutional nets,
  atrous convolution, and fully connected crfs', {\em IEEE transactions on
  pattern analysis and machine intelligence}, {\bf 40}(4),  834--848, (2017).

\bibitem{chen2022deliberated}
Lin Chen, Zhixiang Wei, Xin Jin, Huaian Chen, Miao Zheng, Kai Chen, and Yi~Jin,
  `Deliberated domain bridging for domain adaptive semantic segmentation', {\em
  Advances in Neural Information Processing Systems}, {\bf 35},  15105--15118,
  (2022).

\bibitem{contributors2020mmsegmentation}
MMSegmentation Contributors.
\newblock Mmsegmentation: Openmmlab semantic segmentation toolbox and
  benchmark, 2020.

\bibitem{cordts2016cityscapes}
Marius Cordts, Mohamed Omran, Sebastian Ramos, Timo Rehfeld, Markus Enzweiler,
  Rodrigo Benenson, Uwe Franke, Stefan Roth, and Bernt Schiele, `The cityscapes
  dataset for semantic urban scene understanding', in {\em Proceedings of the
  IEEE conference on computer vision and pattern recognition}, pp. 3213--3223,
  (2016).

\bibitem{dai2018dark}
Dengxin Dai and Luc Van~Gool, `Dark model adaptation: Semantic image
  segmentation from daytime to nighttime', in {\em 2018 21st International
  Conference on Intelligent Transportation Systems (ITSC)}, pp. 3819--3824.
  IEEE, (2018).

\bibitem{fu2019dual}
Jun Fu, Jing Liu, Haijie Tian, Yong Li, Yongjun Bao, Zhiwei Fang, and Hanqing
  Lu, `Dual attention network for scene segmentation', in {\em Proceedings of
  the IEEE/CVF conference on computer vision and pattern recognition}, pp.
  3146--3154, (2019).

\bibitem{gao2022cross}
Huan Gao, Jichang Guo, Guoli Wang, and Qian Zhang, `Cross-domain correlation
  distillation for unsupervised domain adaptation in nighttime semantic
  segmentation', in {\em Proceedings of the IEEE/CVF Conference on Computer
  Vision and Pattern Recognition}, pp. 9913--9923, (2022).

\bibitem{gao2021dsp}
Li~Gao, Jing Zhang, Lefei Zhang, and Dacheng Tao, `Dsp: Dual soft-paste for
  unsupervised domain adaptive semantic segmentation', in {\em Proceedings of
  the 29th ACM International Conference on Multimedia}, pp. 2825--2833, (2021).

\bibitem{hoyer2022daformer}
Lukas Hoyer, Dengxin Dai, and Luc Van~Gool, `Daformer: Improving network
  architectures and training strategies for domain-adaptive semantic
  segmentation', in {\em Proceedings of the IEEE/CVF Conference on Computer
  Vision and Pattern Recognition}, pp. 9924--9935, (2022).

\bibitem{huang2019ccnet}
Zilong Huang, Xinggang Wang, Lichao Huang, Chang Huang, Yunchao Wei, and Wenyu
  Liu, `Ccnet: Criss-cross attention for semantic segmentation', in {\em
  Proceedings of the IEEE/CVF international conference on computer vision}, pp.
  603--612, (2019).

\bibitem{li2019bidirectional}
Yunsheng Li, Lu~Yuan, and Nuno Vasconcelos, `Bidirectional learning for domain
  adaptation of semantic segmentation', in {\em Proceedings of the IEEE/CVF
  Conference on Computer Vision and Pattern Recognition}, pp. 6936--6945,
  (2019).

\bibitem{lin2019adapting}
Yong-Xiang Lin, Daniel~Stanley Tan, Wen-Huang Cheng, and Kai-Lung Hua,
  `Adapting semantic segmentation of urban scenes via mask-aware gated
  discriminator', in {\em 2019 IEEE International Conference on Multimedia and
  Expo (ICME)}, pp. 218--223. IEEE, (2019).

\bibitem{loshchilov2017decoupled}
Ilya Loshchilov and Frank Hutter, `Decoupled weight decay regularization', {\em
  arXiv preprint arXiv:1711.05101}, (2017).

\bibitem{ma2022both}
Xianzheng Ma, Zhixiang Wang, Yacheng Zhan, Yinqiang Zheng, Zheng Wang, Dengxin
  Dai, and Chia-Wen Lin, `Both style and fog matter: Cumulative domain
  adaptation for semantic foggy scene understanding', in {\em Proceedings of
  the IEEE/CVF Conference on Computer Vision and Pattern Recognition}, pp.
  18922--18931, (2022).

\bibitem{olsson2021classmix}
Viktor Olsson, Wilhelm Tranheden, Juliano Pinto, and Lennart Svensson,
  `Classmix: Segmentation-based data augmentation for semi-supervised
  learning', in {\em Proceedings of the IEEE/CVF Winter Conference on
  Applications of Computer Vision}, pp. 1369--1378, (2021).

\bibitem{reinhard2001color}
Erik Reinhard, Michael Adhikhmin, Bruce Gooch, and Peter Shirley, `Color
  transfer between images', {\em IEEE Computer graphics and applications}, {\bf
  21}(5),  34--41, (2001).

\bibitem{sakaridis2019guided}
Christos Sakaridis, Dengxin Dai, and Luc~Van Gool, `Guided curriculum model
  adaptation and uncertainty-aware evaluation for semantic nighttime image
  segmentation', in {\em Proceedings of the IEEE/CVF International Conference
  on Computer Vision}, pp. 7374--7383, (2019).

\bibitem{sakaridis2018model}
Christos Sakaridis, Dengxin Dai, Simon Hecker, and Luc Van~Gool, `Model
  adaptation with synthetic and real data for semantic dense foggy scene
  understanding', in {\em Proceedings of the european conference on computer
  vision (ECCV)}, pp. 687--704, (2018).

\bibitem{sakaridis2019semantic}
Christos Sakaridis, Dengxin Dai, and Luc Van~Gool, `Semantic nighttime image
  segmentation with synthetic stylized data, gradual adaptation and
  uncertainty-aware evaluation', {\em arXiv preprint arXiv:1901.05946}, {\bf
  2}, (2019).

\bibitem{sakaridis2020map}
Christos Sakaridis, Dengxin Dai, and Luc Van~Gool, `Map-guided curriculum
  domain adaptation and uncertainty-aware evaluation for semantic nighttime
  image segmentation', {\em IEEE Transactions on Pattern Analysis and Machine
  Intelligence}, {\bf 44}(6),  3139--3153, (2020).

\bibitem{sakaridis2021acdc}
Christos Sakaridis, Dengxin Dai, and Luc Van~Gool, `Acdc: The adverse
  conditions dataset with correspondences for semantic driving scene
  understanding', in {\em Proceedings of the IEEE/CVF International Conference
  on Computer Vision}, pp. 10765--10775, (2021).

\bibitem{tarvainen2017mean}
Antti Tarvainen and Harri Valpola, `Mean teachers are better role models:
  Weight-averaged consistency targets improve semi-supervised deep learning
  results', {\em Advances in neural information processing systems}, {\bf 30},
  (2017).

\bibitem{tranheden2021dacs}
Wilhelm Tranheden, Viktor Olsson, Juliano Pinto, and Lennart Svensson, `Dacs:
  Domain adaptation via cross-domain mixed sampling', in {\em Proceedings of
  the IEEE/CVF Winter Conference on Applications of Computer Vision}, pp.
  1379--1389, (2021).

\bibitem{vu2019advent}
Tuan-Hung Vu, Himalaya Jain, Maxime Bucher, Matthieu Cord, and Patrick
  P{\'e}rez, `Advent: Adversarial entropy minimization for domain adaptation in
  semantic segmentation', in {\em Proceedings of the IEEE/CVF Conference on
  Computer Vision and Pattern Recognition}, pp. 2517--2526, (2019).

\bibitem{wang2022metateacher}
Zhenbin Wang, Mao Ye, Xiatian Zhu, Liuhan Peng, Liang Tian, and Yingying Zhu,
  `Metateacher: Coordinating multi-model domain adaptation for medical image
  classification', {\em Advances in Neural Information Processing Systems},
  {\bf 35},  20823--20837, (2022).

\bibitem{wortsman2022model}
Mitchell Wortsman, Gabriel Ilharco, Samir~Ya Gadre, Rebecca Roelofs, Raphael
  Gontijo-Lopes, Ari~S Morcos, Hongseok Namkoong, Ali Farhadi, Yair Carmon,
  Simon Kornblith, et~al., `Model soups: averaging weights of multiple
  fine-tuned models improves accuracy without increasing inference time', in
  {\em International Conference on Machine Learning}, pp. 23965--23998. PMLR,
  (2022).

\bibitem{wu2021dannet}
Xinyi Wu, Zhenyao Wu, Hao Guo, Lili Ju, and Song Wang, `Dannet: A one-stage
  domain adaptation network for unsupervised nighttime semantic segmentation',
  in {\em Proceedings of the IEEE/CVF Conference on Computer Vision and Pattern
  Recognition}, pp. 15769--15778, (2021).

\bibitem{wu2021one}
Xinyi Wu, Zhenyao Wu, Lili Ju, and Song Wang, `A one-stage domain adaptation
  network with image alignment for unsupervised nighttime semantic
  segmentation', {\em IEEE Transactions on Pattern Analysis and Machine
  Intelligence}, {\bf 45}(1),  58--72, (2021).

\bibitem{xie2021segformer}
Enze Xie, Wenhai Wang, Zhiding Yu, Anima Anandkumar, Jose~M Alvarez, and Ping
  Luo, `Segformer: Simple and efficient design for semantic segmentation with
  transformers', {\em Advances in Neural Information Processing Systems}, {\bf
  34},  12077--12090, (2021).

\bibitem{xu2021cdada}
Qi~Xu, Yinan Ma, Jing Wu, Chengnian Long, and Xiaolin Huang, `Cdada: A
  curriculum domain adaptation for nighttime semantic segmentation', in {\em
  Proceedings of the IEEE/CVF International Conference on Computer Vision}, pp.
  2962--2971, (2021).

\bibitem{xu2019self}
Yonghao Xu, Bo~Du, Lefei Zhang, Qian Zhang, Guoli Wang, and Liangpei Zhang,
  `Self-ensembling attention networks: Addressing domain shift for semantic
  segmentation', in {\em Proceedings of the AAAI Conference on Artificial
  Intelligence}, volume~33, pp. 5581--5588, (2019).

\bibitem{yang2020fda}
Yanchao Yang and Stefano Soatto, `Fda: Fourier domain adaptation for semantic
  segmentation', in {\em Proceedings of the IEEE/CVF Conference on Computer
  Vision and Pattern Recognition}, pp. 4085--4095, (2020).

\bibitem{yun2019cutmix}
Sangdoo Yun, Dongyoon Han, Seong~Joon Oh, Sanghyuk Chun, Junsuk Choe, and
  Youngjoon Yoo, `Cutmix: Regularization strategy to train strong classifiers
  with localizable features', in {\em Proceedings of the IEEE/CVF international
  conference on computer vision}, pp. 6023--6032, (2019).

\bibitem{zhou2022uncertainty}
Qianyu Zhou, Zhengyang Feng, Qiqi Gu, Guangliang Cheng, Xuequan Lu, Jianping
  Shi, and Lizhuang Ma, `Uncertainty-aware consistency regularization for
  cross-domain semantic segmentation', {\em Computer Vision and Image
  Understanding}, {\bf 221},  103448, (2022).

\bibitem{zhou2022context}
Qianyu Zhou, Zhengyang Feng, Qiqi Gu, Jiangmiao Pang, Guangliang Cheng, Xuequan
  Lu, Jianping Shi, and Lizhuang Ma, `Context-aware mixup for domain adaptive
  semantic segmentation', {\em IEEE Transactions on Circuits and Systems for
  Video Technology}, (2022).

\bibitem{zhu2017unpaired}
Jun-Yan Zhu, Taesung Park, Phillip Isola, and Alexei~A Efros, `Unpaired
  image-to-image translation using cycle-consistent adversarial networks', in
  {\em Proceedings of the IEEE international conference on computer vision},
  pp. 2223--2232, (2017).

\bibitem{zou2018unsupervised}
Yang Zou, Zhiding Yu, BVK Kumar, and Jinsong Wang, `Unsupervised domain
  adaptation for semantic segmentation via class-balanced self-training', in
  {\em Proceedings of the European conference on computer vision (ECCV)}, pp.
  289--305, (2018).

\bibitem{zou2019confidence}
Yang Zou, Zhiding Yu, Xiaofeng Liu, BVK Kumar, and Jinsong Wang, `Confidence
  regularized self-training', in {\em Proceedings of the IEEE/CVF International
  Conference on Computer Vision}, pp. 5982--5991, (2019).

\end{thebibliography}

\end{document}